\documentclass[]{ceurart} 
\usepackage{listings}
\usepackage{amsmath}
\usepackage{amssymb}
\usepackage{booktabs}
\usepackage{xcolor}
\usepackage{enumitem}
\usepackage{float}

\begin{document}

\copyrightyear{2026}
\copyrightclause{Copyright for this paper by its authors.
  Use permitted under Creative Commons License Attribution 4.0
  International (CC BY 4.0).}
\conference{Late-breaking work, Demos and Doctoral Consortium, colocated with the 4th World Conference on eXplainable Artificial Intelligence: July 01--03, 2026, Fortaleza, Brazil}

\title{Quantifying Cross-Modal Interactions in Multimodal Glioma Survival Prediction via InterSHAP: Evidence for Additive Signal Integration}

\author[1]{Iain Swift}[%
email=iain.swift@mymtu.ie,
]
\cormark[1]

\author[1]{Jing Hua Ye}[%
email=JingHua.Ye@mtu.ie,
]

\author[1]{Ruair\'{i} O'Reilly}[%
email=Ruairi.OReilly@mtu.ie,
]

\address[1]{Department of Computer Science, Munster Technological University, Cork, Ireland}

\cortext[1]{Corresponding author.}

\begin{abstract}
Multimodal deep learning for cancer prognosis is commonly assumed to benefit from synergistic cross-modal interactions, yet this assumption has not been directly tested in survival prediction settings. This work adapts InterSHAP, a Shapley interaction index-based metric, from classification to Cox proportional hazards models and applies it to quantify cross-modal interactions in glioma survival prediction. Using TCGA-GBM and TCGA-LGG data ($n{=}575$), we evaluate four fusion architectures combining whole-slide image (WSI) and RNA-seq features. Our central finding is an inverse relationship between predictive performance and measured interaction: architectures achieving superior discrimination (C-index 0.64$\to$0.82) exhibit equivalent or lower cross-modal interaction (4.8\%$\to$3.0\%). Variance decomposition reveals stable additive contributions across all architectures (WSI${\approx}$40\%, RNA${\approx}$55\%, Interaction${\approx}$4\%), indicating that performance gains arise from complementary signal aggregation rather than learned synergy. These findings provide a practical model auditing tool for comparing fusion strategies, reframe the role of architectural complexity in multimodal fusion, and have implications for privacy-preserving federated deployment.
\end{abstract}

\begin{keywords}
Multimodal learning \sep
Explainability \sep
Shapley interaction index \sep
Glioma \sep
Survival prediction \sep
Cross-modal interactions \sep
Model auditing
\end{keywords}

\maketitle

\section{Introduction}
\label{sec:intro}

Gliomas are the most common primary malignant brain tumours, graded 1--4 by the WHO~\cite{Bakas2018}. Glioblastoma (GBM, grade~4) carries median survival of approximately 15~months; lower-grade gliomas (LGG, grades~2--3) have more variable prognosis. Even within identical histological grades, survival varies substantially due to molecular and cellular heterogeneity~\cite{Zheng2023}. This has motivated multimodal deep learning approaches integrating histopathology and genomics to improve prognostic accuracy~\cite{Steyaert2023, Mobadersany2018}.

\textit{The Interaction Learning Assumption.}
Performance improvements from multimodal fusion are commonly attributed to \emph{interaction learning}: the hypothesis that models discover predictive patterns depending on both modalities jointly~\cite{Steyaert2023}. For example, a model might learn that a specific histological pattern combined with a particular gene expression signature indicates poor prognosis, even when neither alone is predictive. However, this interpretation rests on a logical gap: \emph{performance improvement does not necessarily imply interaction learning}. A model combining independent, complementary signals achieves better predictions without discovering such emergent cross-modal phenotypes.

Distinguishing \emph{complementarity} (additive improvement from independent signals) from \emph{synergy} (non-additive improvement requiring both modalities simultaneously) requires direct measurement of the interaction component. Yet no prior work has rigorously quantified cross-modal interactions in survival prediction.

This distinction has practical consequences. If models operate additively, simpler architectures suffice, modality-specific explanations remain valid for regulatory compliance, and federated deployment is simplified since institutions can train unimodal encoders independently~\cite{McMahan2017}. If models operate synergistically, more complex fusion mechanisms and joint explainability approaches are required.

\textit{Contributions.} This paper makes two contributions. \textit{Methodological:} We adapt InterSHAP~\cite{Wenderoth2025}, a Shapley interaction index-based~\cite{Grabisch1999} metric, from classification to Cox survival models, validated on synthetic data with known interaction patterns. \textit{Empirical:} We demonstrate an inverse relationship between predictive performance and measured interaction across four fusion architectures applied to glioma survival prediction, with stable variance decomposition (WSI${\approx}$40\%, RNA${\approx}$55\%, Interaction${\approx}$4\%).

InterSHAP is positioned as a \emph{model auditing tool}. While patient-level variation in InterSHAP correlates with survival, this reflects model-internal structure rather than independent prognostic evidence, since InterSHAP values are deterministic functions of the model's own predictions~\cite{Kaufman2012, Rudin2019}.

\section{Related Work}
\label{sec:related}

\textit{Multimodal Fusion for Glioma Prognosis.}
Multimodal integration for cancer prognosis has evolved from feature concatenation~\cite{Mobadersany2018} to attention mechanisms~\cite{Ilse2018, Lu2021}. Steyaert et al.~\cite{Steyaert2023} compared early, late, and joint fusion of FFPE histopathology and gene expression data (RNA-seq and microarray) on adult ($n{=}783$) and pediatric ($n{=}305$) glioma cohorts. Their adult early fusion model achieved a composite score (average of concordance index and $1{-}$IBS) of 0.831$\pm$0.022 on cross-validation. They attributed gains to synergistic learning but provided no direct interaction measurement. Notably, Steyaert et al.\ used gradient-based backpropagation for pathway-level interpretation of their RNA encoder, visualised with SHAP summary plots, identifying biologically relevant cancer pathways including \emph{RAS} and \emph{NTRK} signalling.

\textit{Shapley Interaction Indices.}
SHAP~\cite{Lundberg2017} provides game-theoretic feature attribution. The Shapley Interaction Index (SII), axiomatised by Grabisch and Roubens~\cite{Grabisch1999}, extends this to pairwise interactions. For features $i,j$ in model $f$:
\begin{equation}
\psi_{ij}(f) = \!\!\!\sum_{S \subseteq N \setminus \{i,j\}} \!\!\!\frac{|S|!(|N|{-}|S|{-}2)!}{(|N|{-}1)!}\, \Delta_{ij}(S, f)
\label{eq:sii}
\end{equation}
where $\Delta_{ij}(S, f) = f(S \cup \{i,j\}) - f(S \cup \{i\}) - f(S \cup \{j\}) + f(S)$ is the discrete second derivative, equalling zero for additive models. InterSHAP~\cite{Wenderoth2025} applies SII at the modality level, satisfying efficiency, symmetry, dummy, and additivity axioms. Unlike attention maps or gradient methods, it provides architecture-agnostic quantification grounded in cooperative game theory.

\textit{Limits of Post-Hoc Explanation.}
Lipton~\cite{Lipton2018} highlighted interpretability ambiguities; Rudin~\cite{Rudin2019} cautioned against post-hoc explanations in high-stakes decisions; Kaufman et al.~\cite{Kaufman2012} formalised target leakage risks. These critiques inform our positioning: InterSHAP values are model-derived quantities and should not be treated as independent clinical variables.

\section{Methodology}
\label{sec:method}

\subsection{Dataset and Preprocessing}

Data were drawn from the TCGA-GBM~\cite{TCIA_GBM2016} and TCGA-LGG~\cite{TCIA_LGG2016} cohorts. The combined TCGA-GBMLGG collection was quality-filtered for patients with matched whole-slide images (WSI), RNA-seq profiles, and survival outcomes, yielding $n{=}575$. This is smaller than the 783 in Steyaert et al.~\cite{Steyaert2023}, who additionally included 199~GBM patients with microarray expression data; our stricter filtering requires RNA-seq specifically. Preprocessing replicates their pipeline.

The cohort includes 195 death events (34\%), median follow-up 14.3~months (censored), GBM ($n{=}67$) and LGG ($n{=}508$). Stratification uses histological diagnosis rather than molecular classification (IDH mutation status), a limitation discussed in Section~\ref{sec:discuss}. The Karnofsky Performance Status (KPS), which quantifies functional ability on a 0--100 scale, was available for a subset and is used in confounding analysis.

\textit{Histopathology:} WSIs were processed at 20$\times$ magnification into non-overlapping 224$\times$224 patches. ResNet-50 (ImageNet-pretrained) extracted 2,048-dimensional features per patch, averaged to slide-level embeddings~\cite{Steyaert2023}.

\textit{Transcriptomics:} RNA-Seq by Expectation-Maximization (RSEM) normalised RNA-seq profiles were log-transformed, z-score normalised, and filtered to genes with ${>}$80\% non-zero expression (12,778 features). A three-layer MLP encoder (12,778$\to$4,096$\to$2,048, dropout 0.5, ReLU) produced embeddings matching WSI dimensionality.

\textit{Data splits:} 80/20\% train/test stratified by outcome; 10-fold cross-validation for robustness.

\subsection{Fusion Architectures}

\begin{figure}[tbp]
\centering
\includegraphics[width=0.72\textwidth]{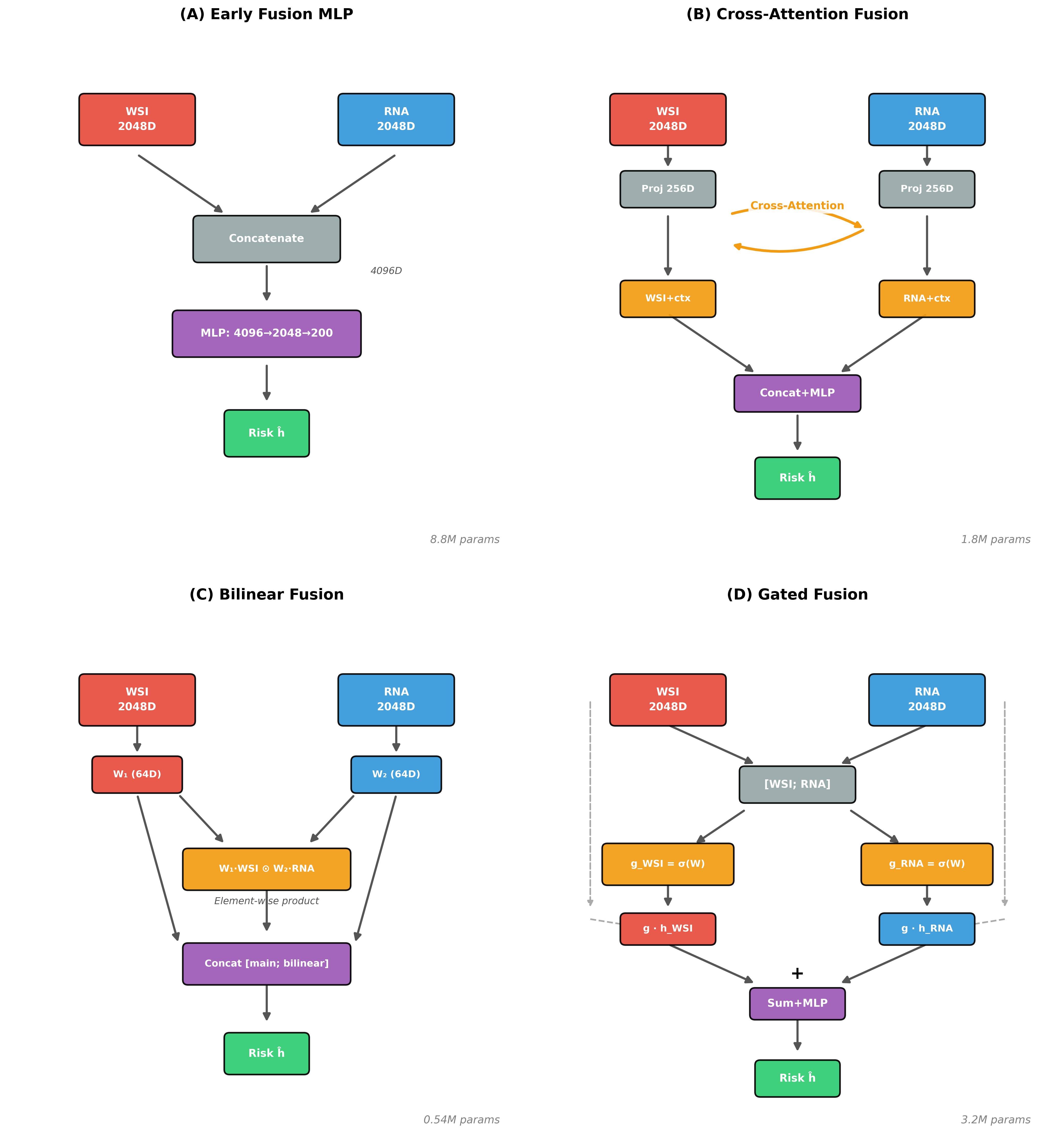}
\caption{Fusion architectures evaluated. (A)~Early Fusion MLP (8.8M params). (B)~Cross-Attention (1.8M params). (C)~Bilinear Fusion (0.54M params). (D)~Gated Fusion (3.2M params).}
\label{fig:architectures}
\end{figure}

We evaluate four strategies of increasing complexity (Figure~\ref{fig:architectures}), each receiving the same 2,048D WSI and RNA embeddings.
\textit{Early Fusion MLP (Baseline):} Concatenation followed by MLP (4096$\to$2048$\to$200$\to$1), ReLU, dropout (0.25), replicating~\cite{Steyaert2023}. Learns interactions implicitly through hidden layers (8.8M params).
\textit{Cross-Attention Fusion:} Bidirectional attention: $A_{\text{WSI}\to\text{RNA}} = \text{softmax}(Q_{\text{WSI}} K_{\text{RNA}}^T / \sqrt{d})\, V_{\text{RNA}}$. Explicitly models cross-modal dependencies (1.8M params).
\textit{Bilinear Fusion:} Low-rank multiplicative interaction: $z = (W_1 \cdot \text{RNA}) \odot (W_2 \cdot \text{WSI})$, $W_i \in \mathbb{R}^{64 \times 2048}$. Forces outer-product interactions (0.54M params).
\textit{Gated Fusion:} Dynamic weighting: $\alpha = \sigma(W_g[\text{RNA}; \text{WSI}])$; $z = \alpha \odot f(\text{RNA}) + (1{-}\alpha) \odot g(\text{WSI})$ (3.2M params).
All models use Cox partial likelihood loss~\cite{Cox1972}, Adam (lr $10^{-4}$, weight decay $10^{-5}$), batch size 128, and early stopping on validation C-index (patience 20). The concordance index (C-index) measures the probability that predicted risk correctly orders a random patient pair by survival time. Results are mean $\pm$ SD from 5-seed training.

\subsection{InterSHAP for Cox Survival Models}
\label{subsec:intershap_method}

InterSHAP~\cite{Wenderoth2025} was adapted from classification with three design choices:

\textit{Output space:} Shapley values computed on the \textit{log-risk score} $\hat{h}$ (raw network output) rather than $\exp(\hat{h})$, since the exponential introduces multiplicative structure conflating learned interactions with Cox formulation artifacts.

\textit{Coalition evaluation:} Four coalitions per patient: $v(\emptyset)$ (both masked), $v(\{\text{WSI}\})$, $v(\{\text{RNA}\})$, $v(\{\text{WSI},\text{RNA}\})$ (full model). Missing modalities replaced by dataset-mean embeddings; robustness validated against random-shuffle and zero-imputation.

\textit{Shapley computation:} For $M{=}2$, the main effects and interaction are:
\begin{align}
\phi_{\text{WSI}} &= \tfrac{1}{2}[v(\{\text{WSI}\}) {-} v(\emptyset)] + \tfrac{1}{2}[v(\{\text{WSI,RNA}\}) {-} v(\{\text{RNA}\})] \\
\phi_{\text{RNA}} &= \tfrac{1}{2}[v(\{\text{RNA}\}) {-} v(\emptyset)] + \tfrac{1}{2}[v(\{\text{WSI,RNA}\}) {-} v(\{\text{WSI}\})] \\
\phi_{\text{int}} &= \tfrac{1}{2}[v(\{\text{WSI,RNA}\}) {-} v(\{\text{WSI}\}) {-} v(\{\text{RNA}\}) + v(\emptyset)]
\end{align}
The interaction $\phi_{\text{int}}$ is the discrete second derivative at the modality level: zero when the model is additive, non-zero when joint effects exceed the sum of parts. Global InterSHAP expresses interaction as a percentage:
\begin{equation}
\text{InterSHAP}_{\text{global}} = \frac{\sum_i |\phi_{\text{int}}^{(i)}|}{\sum_i (|\phi_{\text{WSI}}^{(i)}| + |\phi_{\text{RNA}}^{(i)}| + |\phi_{\text{int}}^{(i)}|)} \times 100\%
\end{equation}

\textit{Worked example.} As a hypothetical illustration, consider a model producing log-risk scores $v(\emptyset){=}0.0$, $v(\{\text{WSI}\}){=}0.4$, $v(\{\text{RNA}\}){=}0.6$, $v(\{\text{WSI,RNA}\}){=}1.05$ for a single patient. Then $\phi_{\text{int}} = \tfrac{1}{2}[1.05{-}0.4{-}0.6{+}0.0] = 0.025$, with main effects $\phi_{\text{WSI}}{=}0.425$ and $\phi_{\text{RNA}}{=}0.6$ summing with the interaction to recover the full prediction $1.05$ (efficiency axiom). Interaction here accounts for only $0.025/(0.425{+}0.6{+}0.025) \approx 2.4\%$ of the absolute contribution, indicating near-additive behaviour. \emph{Negative} InterSHAP values arise when the joint prediction is \emph{less} than the sum of the individual contributions, indicating \emph{redundancy} between modalities rather than synergy.

\textit{Computational cost:} Four forward passes $\times$ 575 patients = 2,300 evaluations per architecture (${\approx}$3~min, RTX 3060). Exact for $M{=}2$; scaling to $M{>}2$ requires $2^M$ coalitions (tractable for $M{=}3$, non-trivial for $M{\geq}5$). Sampling-based Shapley approximations may extend applicability to higher-dimensional settings.

\subsection{Validation Protocol}

\textit{Synthetic verification:} Survival data with known patterns (uniqueness: expected 0\%, observed ${<}$1\%; XOR synergy: expected ${\approx}$100\%, observed 99.7\%; redundancy: expected 30--50\%, observed within range) aligned with the original InterSHAP validation~\cite{Wenderoth2025}.

\textit{Late fusion zero-check:} A linear combination of independent risk scores must yield zero interaction by construction; verified: $|\text{InterSHAP}| < 10^{-15}$.

\textit{Modality contribution bounds:} Both modalities contribute ${>}$15\%, ruling out modality collapse (WSI: 38$\pm$22\%, RNA: 62$\pm$22\%).

\textit{Cross-validation stability:} 10-fold CV on Early Fusion MLP: mean InterSHAP 1.05\% $\pm$ 0.07\%, coefficient of variation 6.3\%.

\section{Results}
\label{sec:results}

\subsection{Model Performance and Replication}

All architectures outperformed unimodal baselines (Table~\ref{tab:performance}). Bilinear Fusion achieved the best discrimination (C-index 0.819) and calibration (Brier 0.151) with 16$\times$ fewer parameters than MLP, suggesting architectural efficiency rather than capacity drives improvement.

\begin{table}[h]
\centering
\caption{Performance and cross-modal interaction across architectures. Mean $\pm$ SD from 5-seed training. InterSHAP on held-out test set ($n{=}115$).}
\small
\begin{tabular}{lcccc}
\toprule
Architecture & Params & C-Index & InterSHAP (\%) & Brier@36mo \\
\midrule
Early Fusion MLP & 8.8M & 0.636 $\pm$ 0.02 & 4.82 $\pm$ 4.65 & 0.192 \\
Cross-Attention & 1.8M & 0.814 $\pm$ 0.01 & 3.03 $\pm$ 2.64 & 0.156 \\
Bilinear Fusion & 0.54M & 0.819 $\pm$ 0.01 & 3.72 $\pm$ 3.16 & 0.151 \\
Gated Fusion & 3.2M & 0.807 $\pm$ 0.02 & 4.45 $\pm$ 4.94 & 0.162 \\
\bottomrule
\end{tabular}
\label{tab:performance}
\end{table}

\textit{Replication gap.} The MLP (C-index 0.636) underperforms the composite score of 0.831$\pm$0.022 in~\cite{Steyaert2023}, complicated by differing metrics (C-index vs.\ composite score), cohort size ($n{=}575$ vs.\ 783), and unspecified hyperparameters. A nested 5-fold CV improved our MLP to 0.71. This gap does not affect internal validity: all architectures are compared under identical conditions, and attention models (C-index 0.81--0.82) achieve discrimination comparable to the original.

\subsection{Inverse Performance--Interaction Relationship}

The central finding: \textit{models with higher predictive performance exhibit equivalent or lower cross-modal interaction} (Table~\ref{tab:inverse}).

\begin{table}[h]
\centering
\caption{Performance versus interaction. Better models show lower measured interaction. Bootstrap comparison (1,000 iterations) against MLP baseline.}
\small
\begin{tabular}{lccc}
\toprule
Architecture & C-Index & InterSHAP (\%) & $\Delta$ vs MLP (95\% CI) \\
\midrule
MLP (baseline) & 0.636 & 4.82 & --- \\
Cross-Attention & 0.814 & 3.03 & $-1.79$ ($-3.12, -0.46$)$^{***}$ \\
Bilinear & 0.819 & 3.72 & $-1.10$ ($-2.38, -0.18$)$^{**}$ \\
Gated & 0.807 & 4.45 & $-0.37$ ($-1.89, +1.15$)$^{\text{ns}}$ \\
\bottomrule
\end{tabular}
\begin{flushleft}
\footnotesize{$^{***}p < 0.001$; $^{**}p < 0.01$; $^{\text{ns}}$ not significant ($p{=}0.43$). Spearman $\rho = -0.80$ ($n{=}4$; interpret with caution).}
\end{flushleft}
\label{tab:inverse}
\end{table}

Cross-Attention achieved C-index 0.814 yet exhibited the \emph{lowest} InterSHAP (3.03\%). The MLP, with inferior discrimination, showed the highest (4.82\%). This inverse pattern challenges the assumption that architectural complexity improves predictions by learning cross-modal synergies.

High standard deviations (e.g., 4.82 $\pm$ 4.65\% for MLP) reflect patient-level heterogeneity rather than seed instability; per-seed coefficient of variation was ${<}$15\%.

\subsection{Variance Decomposition}

The contribution of each component was remarkably stable across all four architectures. WSI contributed 42.2$\pm$2.1\% (MLP), 43.2$\pm$1.8\% (Cross-Attention), 36.8$\pm$2.4\% (Bilinear), and 43.7$\pm$2.0\% (Gated). RNA contributed 53.0$\pm$2.3\%, 53.8$\pm$1.9\%, 59.5$\pm$2.6\%, and 51.8$\pm$2.2\% respectively. Interaction contributed 4.8$\pm$0.8\%, 3.0$\pm$0.5\%, 3.7$\pm$0.6\%, and 4.5$\pm$0.9\%. The consistency is striking, particularly for Bilinear Fusion, which explicitly forces multiplicative interaction yet shows only 3.7\%. Whether this additive structure reflects genuine biological independence between histopathology and transcriptomics, or arises from independent encoder pre-training linearising interactions before fusion, cannot be determined from this analysis alone.

\subsection{Robustness Analyses}

\textit{Masking strategy:} Mean imputation (3.03\%) and random shuffle (3.51\%) yielded consistent results for Cross-Attention. Zero imputation inflated InterSHAP to 12.13\% by moving inputs off the data manifold, implying that elevated interaction measurements in other work should be scrutinised for masking artifacts.

\textit{Tumour type stratification:} InterSHAP did not differ between GBM and LGG ($p = 0.86$), confirming additive behaviour across glioma grades. WSI contributed more in GBM (49.7\% vs.\ 37.9\%, $p = 0.04$), consistent with distinctive high-grade histological features (pseudopalisading necrosis, microvascular proliferation).

\textit{Negative results:} Patch-level InterSHAP was computationally infeasible ($>$50,000 forward passes per patient). Gene-set InterSHAP was unstable across seeds. Time-varying InterSHAP at landmarks (12, 24, 36~months) was inconclusive due to reduced sample sizes.

\subsection{Exploratory Observation: Patient-Level InterSHAP and Survival}

Patient-level variation in InterSHAP correlates with outcomes: above-median patients showed worse survival (median 33.6 vs.\ 114.0~months; HR$=$1.96, 95\% CI: 1.76--2.18), persisting after tumour-type adjustment (HR$=$1.42, $p<0.0001$). A monotonic dose-response across quartiles was observed (Q1: 114.0~months, Q4: 16.8~months; trend $p < 10^{-36}$). The effect was concentrated in LGG (HR$=$1.99, $p<0.0001$); GBM ($n{=}67$) showed no significant effect (HR$=$1.06, $p=0.29$), though underpowered given uniformly poor prognosis.

Confounding analysis revealed that high-InterSHAP patients were older (51.8 vs.\ 43.2~years, $p<0.001$), more likely GBM (16.4\% vs.\ 7.3\%, $p=0.001$), and had lower KPS ($p=0.004$). After adjusting for age, sex, tumour type, and KPS, InterSHAP remained significant (HR$=$1.42, 95\% CI: 1.21--1.67), though attenuated from 1.96, indicating partial capture of known prognostic factors.

We emphasise that this reflects \emph{model-internal structure}, not independent prognostic evidence. Following Kaufman et al.~\cite{Kaufman2012}: if $f(x)$ is the model's log-hazard, then InterSHAP$(x) = g(f(x))$ for fixed decomposition $g$, and correlation of $g(f(x))$ with survival is expected, not novel. Attention weights in the Cross-Attention model showed weak correlation with InterSHAP ($r = 0.12$), confirming that these capture different aspects of model behaviour. Establishing biological meaning would require external validation, comparison with IDH/MGMT markers, and cross-model invariance, none of which are met here.

\section{Discussion}
\label{sec:discuss}

\textit{Interpreting the Inverse Relationship.}
We expected attention-based models to show \emph{higher} interaction, given their explicit cross-modal query-key-value mechanisms. The inverse pattern (better models, less interaction) suggests three explanations: (1)~\emph{Efficient signal aggregation:} attention improves unimodal encoding rather than cross-modal fusion. Evidence includes 28\% C-index improvement with reduced InterSHAP, stable decomposition, and Bilinear Fusion showing only 3.7\% despite forced multiplicative terms. (2)~\emph{Interaction suppression:} attention mechanisms may filter spurious cross-modal correlations that do not generalise. (3)~\emph{Biological independence:} histopathology (cellular architecture) and transcriptomics (molecular pathways) capture orthogonal aspects of glioma biology at the embedding level.

Efficient signal aggregation appears most plausible. Interaction suppression is unlikely: Bilinear Fusion cannot suppress multiplicative terms by design, yet shows low InterSHAP. Several observations also argue against underfitting as an alternative explanation: attention models achieve C-index 0.82, demonstrating capacity to learn complex patterns; systematic InterSHAP differences exist between architectures (MLP: 4.8\% vs.\ Cross-Attention: 3.0\%); and the inverse relationship itself is inconsistent with underfitting, which would predict better models learn \emph{more} interaction, not less. A further possibility is that interactions are absorbed inside the encoders themselves: the ResNet-50 and MLP encoders are pre-trained independently, potentially linearising cross-modal relationships before they reach the fusion stage. Whether end-to-end joint training of encoders would expose higher measurable interactions is an open empirical question for future work.

\textit{Biological Interpretation.}
We hypothesise that elevated patient-level InterSHAP may reflect signal discordance: cases where histological and molecular features provide conflicting prognostic information. In most patients, aggressive morphology may correspond to aggressive molecular signatures, yielding low interaction through concordant additive signals. Patients with elevated InterSHAP may exhibit: intratumoral heterogeneity (different tumour regions with different molecular-morphological relationships), rapid molecular evolution not yet manifest morphologically, or microenvironment complexity from immune infiltration creating non-additive histology-transcriptome relationships. The concentration of the prognostic effect in LGG (HR$=$1.99) rather than GBM (HR$=$1.06) is consistent with this hypothesis: LGG has greater molecular heterogeneity and longer disease trajectories allowing more opportunity for discordance. This remains hypothesis-generating and requires validation with IDH-stratified cohorts.

\textit{Relevance to XAI.}
InterSHAP provides an architecture-agnostic, axiomatically grounded tool for auditing multimodal model behaviour. Unlike attention maps (architecture-specific) or gradient methods (local sensitivity), InterSHAP quantifies the actual contribution of interaction to predictions through game-theoretic decomposition, making it suitable for regulatory documentation (e.g., EU AI Act). If models operate additively, modality-specific explanations remain valid independently, simplifying both clinical communication and federated deployment~\cite{McMahan2017}, where modality encoders can train separately with only scalar risk scores shared.

\subsection{Limitations}

The TCGA cohort ($n{=}575$) is modest with no external validation; larger datasets might reveal learnable interactions. InterSHAP operates on 2,048D learned embeddings, so interactions at lower levels may be absorbed during encoding; whether independent encoder pre-training linearises interactions is an open question. Stratification by GBM/LGG rather than IDH status is significant, as 10--20\% of LGG may be IDH-wildtype. Scalability is limited by the $2^M$ coalition requirement, though $M{=}3$ (adding MRI) is tractable and planned. InterSHAP is a single global score per patient; time-varying analyses were inconclusive. The MLP replication gap (Section~\ref{sec:results}) means absolute InterSHAP values should not be compared across studies.

\section{Conclusion}
\label{sec:conclude}

We adapted InterSHAP for Cox survival models and quantified cross-modal interactions in multimodal glioma prediction. Two principal findings emerge: (1)~an inverse performance--interaction relationship where architectures achieving superior discrimination (C-index 0.64$\to$0.82) exhibit equivalent or lower interaction than simple concatenation; and (2)~stable additive decomposition (WSI: 37--44\%, RNA: 52--60\%, Interaction: 3--5\%) across four architectures of varying complexity.

These findings have several practical implications. First, simple fusion architectures may be sufficient for glioma survival prediction, as architectural complexity primarily improves unimodal signal extraction rather than cross-modal fusion. Second, performance improvements from multimodal models should not be automatically interpreted as evidence of interaction learning, since complementarity and synergy are distinct phenomena requiring separate measurement. Third, the additive structure simplifies both clinical interpretation (modality-specific explanations remain valid independently) and privacy-preserving federated deployment (unimodal encoders can be trained at separate institutions).

Future work will extend InterSHAP to three modalities (MRI from BraTS-TCGA), evaluate IDH-stratified interaction patterns within LGG, and test whether end-to-end joint encoder training produces higher measured interactions.\footnote{Code: \url{https://github.com/iainswift/intershap-glioma}}

\begin{acknowledgments}
The authors thank the Department of Computer Science at Munster Technological University for supporting this research.
\end{acknowledgments}

\section*{Declaration on Generative AI}
During the preparation of this work, the author(s) used Claude (Anthropic) for grammar checking. The author(s) take(s) full responsibility for the publication's content.


\end{document}